\documentclass[conference]{IEEEtran}

\usepackage{graphicx}
\usepackage[usenames,dvipsnames]{xcolor}
\usepackage{url}  
\usepackage{psfrag}  
\usepackage{amsthm}
\usepackage{amsmath}
\usepackage{amsfonts}
\usepackage{graphicx}
\usepackage{wrapfig}
\usepackage{color}
\usepackage[lined, ruled, boxed, linesnumbered]{algorithm2e}

\newtheorem*{define}{Definition}	

\begin{document}
\IEEEoverridecommandlockouts

\IEEEpubid{\makebox[\columnwidth]{978-1-4799-7560-0/15/\$31 \copyright 2015 IEEE \hfill} \hspace{\columnsep}\makebox[\columnwidth]{ }}

\title{Evolving Non-linear Stacking Ensembles\\
		for Prediction of Go Player Attributes}

\author{\IEEEauthorblockN{Josef~Moud\v{r}\'{i}k}
\IEEEauthorblockA{ Charles University in Prague\\ Faculty of Mathematics and Physics\\
Malostransk\'e n\'am\v{e}st\'\i~25\\
Prague, Czech Republic\\
email: j.moudrik@gmail.com}%
\and\IEEEauthorblockN{Roman Neruda}%
\IEEEauthorblockA{Institute of Computer Science\\ Academy of Sciences of the Czech Republic\\
Pod Vod\'{a}renskou v\v{e}\v{z}\'\i~2\\
 Prague, Czech Republic\\
email: roman@cs.cas.cz}
}

\maketitle

\begin{abstract}

The paper presents an application of non-linear 
stacking ensembles for prediction of Go player
attributes. An evolutionary algorithm is used
to form a diverse ensemble of base learners,
which are then aggregated by a stacking ensemble.
This methodology allows for an efficient prediction
of different attributes of Go players from sets of
their games. These attributes can be fairly general,
in this work, we used the strength and style of
the players.

\end{abstract}


\section{Introduction}
The field of computer Go is primarily focused on the problem
of creating a~program to play the game by finding the best move from a~given
board position~\cite{GellySilver2008}. We focus on analyzing existing game
records with the aim of helping humans to play and understand the game better
instead. In our previous work~\cite{Moudrik15}, we have presented 
a way to extract information rich features from sets of Go players' games.

This paper presents machine-learning methodology we have devised to fully
utilize these extracted features. We have used stacking ensembles with
non-linear second-level learner. Both the members of the ensemble and the
second-level learner are chosen by a genetic algorithm.
The resulting model performs better than any single base-learner
(see Section~\ref{sec:learn}) on its own, and also better than the best
hand-tuned ensemble we have been able to come up with.

As the methods of this work are mainly domain-independent, we only
give a brief introduction to Go-related specifics.
Go is a~two-player full-information board game played
on a~square grid (usually $19\times19$ lines) with black and white
stones; the goal of the game is to surround territory and
capture enemy stones. Different players tend to choose different
strategies to achieve this, in Go terminology this is captured
by the notion of playing style. Obviously, the players also vary in
their proficiency, to which we refer to as \emph{strength}. All these
player attributes (various axes of style, and strength) can be mapped
on a subset of real numbers~\cite{Moudrik13}. In this paper, we 
use thusly defined regression domains as black-box datasets.

This paper is organized as follows. Section~\ref{sec:rel} gives
overview of related work.
Section~\ref{sec:learn} presents the learners
(Sections \ref{mach:base_learners} and~\ref{mach:ensemble_learners})
and the genetic algorithm (Section~\ref{mach:ga}).
Section~\ref{mach:cmp} discusses performance measures used
to compare various learners. Experiments and results are shown
in Section~\ref{sec:expe}. Section~\ref{sec:disc} discusses
applications and future work.

\section{Related Work}
\label{sec:rel}

Ensemble approaches to machine-learning have been in researchers'
attention for over two decades. During the time, various approaches
appeared. Some of them train one model on differently sampled data
\cite{hansen1990neural}, as is the case of bagging~\cite{breimanbag96}
and the related random forests algorithm~\cite{breiman01}.
The schemes for combining such models in a voting-like manner
seems to be well understood~\cite{kittler1998combining}. 
For example, in neural networks, these ideas have also recently been
re-introduced in the form of a dropout~\cite{hinton2012improving}.

Another approach to combine different models is boosting~\cite{boosting},
where a (presumably weak) model is iteratively trained to specialize
on hard instances. Stacking~\cite{wolpert92} on the
other hand, uses a two-layered approach, where model on the second
level learns to correct for mistakes that first level learners
make. For classification, various ways of forming the features from
the first level prediction have been proposed
(\cite{ting1999issues,chan1996extensible}), multi-response linear regression
has been found to work well for second level learner.
For regression task such as ours, simple linear second level models
have been proposed by Breiman~\cite{breiman96}. We are not aware
of any use of non-linear models for second level predictors like
we use in this work.

Prediction of Go player attributes has until recently been limited
to pre-defined questionnaires and simple methods \cite{style:baduk},
\cite{senseis:which_pro}. Universal approach to the problem has been
introduced by our previous work \cite{Moudrik15} and \cite{Moudrik13}.

\section{Learners}
\label{sec:learn}

This section presents the machine learning framework we have used. The basic
methods are well-documented in literature, so we only give a very brief overview here.

Suppose we have a set of data
$$ Tr = \{ (x_1, y_1), \ldots, (x_N, y_N) \}, \forall i: x_i \in \mathbb{R}^p, y_i \in \mathbb{R},$$
and we want to find a function $r$ which is able to predict the value $y_i$
from $x_i$ with a reasonable accuracy and can generalize this dependency to unseen pairs.
The machine learning methods presented here are regarded as
\emph{learners}. For a given data $Tr$, the \emph{learner} should output a
\emph{regression function} which performs the regression of the dependent variable,
as learned from the data.

Of course, some \emph{regression functions} perform better than others.
Mainly, this is because each \emph{learner} has different (inherent)
assumptions about the form of the function it is fitting; we call
this the \emph{inductive bias} (of the \emph{learner} and the underlying model).
Often, we deal with data where the underlying dependency and properties
of the data are unknown, so it is hard to say whether
assumptions of a particular model are right. To overcome this problem,
usually a bunch of models is tried and the best one is chosen (according to some criterion).

Another approach, the one we use in this work, is not to choose the best,
but rather try to combine the different models to create one higher-level
method.  Because different methods have different biases, they might be able to
capture different dependencies in the data. If we combined the methods (%
\emph{base learners})
usefully, we could get better performance than with the ``use the best learner''
approach, we understand this as \emph{ensemble learning}.


\subsection{Base Learners}
\label{mach:base_learners}

\subsubsection{Mean Regression}
\label{mach:mean}
is a very simple method, which we use as a reference for
comparing performances of other learners. It simply outputs the mean of the $y$'s
in the training set, and it is thus constant regardless of the input $x$.

$$ mean(x) = \frac{1}{|Tr|} \sum_{(x', y') \in Tr}{y'} $$

\subsubsection{Artificial Neural Networks}
are a standard technique used for function approximation. 
The network is composed of simple computational units which are organized
in a layered topology, as described e.g. in a monograph \cite{haykin_nn}.
We have used a simple feedforward neural network with 20~hidden units in one
hidden layer. The neurons have standard sigmoidal activation function and
the network is trained using the RPROP algorithm~\cite{Riedmiller1993} for at most
100~iterations (or until the error is smaller than 0.001). In both datasets 
used, the domain of the particular target variable (strength, style) was
linearly rescaled to $\langle -1,1 \rangle$ prior to learning. Similarly,
predicted outputs were rescaled back by the inverse mapping.

\subsubsection{$k$-Nearest Neighbor Regression}

is another commonly used machine learning algorithm~\cite{CoverHart1967}. The assumption
of this model is that we can deduce the dependent $y$ by
looking at vectors from the feature space that are close to the $x$.

\begin{define}
    For a fixed $k$ and $x$, let the $Nb = \{x'_1, \ldots, x'_k \}$ denote a set of $k$ closest
    vectors to $x$ from the $T$ with respect to some metric $\delta$; 
    let $D$ be a vector of distances, such that $D_i = \delta(x, x'_i)$;
    for each $x'_i$, let $y'_i$ be the associated dependent variable from the training set $T$.
\end{define}

For a given $x$, the idea is to find the nearest $k$ vectors (the $Nb$ set)
from the training set,
and then estimate the dependent variable $y$ from the associated $y'_1, \ldots, y'_k$.

In this work, we have used the Manhattan ($p$-1) and Euclidean ($p$-2) distances as $\delta$.
To infer the $y$, we define the model to be:
$$ y = \frac{\sum_{i=1}^k{w(D_i) y'_i}}{\sum_{i=1}^k{w(D_i)}}, $$
for some weighting function $w$. We have used the inverse of
the distance between $x$ and the particular neighbor instance:
$$ w(D_i) = 1 / D_i^\alpha,$$
where $\alpha$ is a parameter specifying the effect of increasing distance.
When $\alpha$ is equal to zero, we obtain the averaging scheme, where
the weights do not depend on the distance $D_i$---all the $k$ neighbors are
valued equally. With increasing $\alpha$, the $x'_i$ instances closer to $x$
are preferred over more distant neighbors. When $\alpha$ goes to infinity,
the method essentially becomes one-nearest neighbor.

\subsubsection{PLS}
The family of \emph{partial least squares} (\emph{PLS}) methods
assumes that the observed variables
can be modelled by means of a few latent variables (their number is specified
by a parameter $l$).
The method projects the data onto this latent model in a way that minimizes error.
The process is somewhat similar to Principal Component Regression.

For a good overview, see the work \cite{pls}.

\subsubsection{Random Forests}
utilize an ensemble of tree learners
to predict the dependent value
\cite{breiman01}, \cite{breiman84}. Although this 
model is an ensemble itself, we treat is as a black box in this work.

Each tree from the forest (of size $N$) is trained on an independently chosen
subset of training data, exactly as the bagging in Section~\ref{mach:bagging} does.
See the Breiman's paper \cite{breiman01} for details.

\begin{table}
\begin{center}
\caption{Base Learners and their Settings.}
\label{tab:base_learners}
\begin{tabular}{|c|c|}
\hline
\textbf{Base learner} & \textbf{Settings} \\
\hline
Mean regression & --- \\
PLS regression & $l \in \{2, \ldots, 10\}$ \\

$k$-nearest neighbors & $k \in \{10, 20, \ldots, 60\}$, $\alpha \in \{10, 20\}$,\\
				  &	$\delta \in \{\text{Manhattan},\text{Euclidean}\}$, all combinations. \\ 
Random Forests & $N \in \{5, 10, 25, 50, 100, 200\}$\\

Neural network & $max \in \{50, 100,200, 500\}$ iterations \\
			   & $\epsilon \in \{0.001, 0.005\}$, 1 layer with\\
			&number of neurons $\in \{10, 20\}$, all combinations.\\
& Symmetric sigmoid activation function.\\

Bagged Neural & For ensemble sizes of $\in \{ 20, 40, 60 \}$, each\\
Networks & Neural network (from right above) was tested. \\


\hline
\end{tabular}
\end{center}
\end{table}

\subsection{Ensemble Learners}
\label{mach:ensemble_learners}

\subsubsection{Bagging}
\label{mach:bagging}
is a simple ensemble method introduced in~\cite{breimanbag96}.
The idea in bagging is to train a particular base learner $bl$ on
differently sampled data and aggregate the results. The method has one
parameter $t$ which specifies the number of data samples. Each of them
is made by randomly choosing $|Tr|$ elements from training set $Tr$
\emph{with repetition}. The base learner $bl$ is trained on each of
these samples. The regression simply averages results from the $t$
resulting models.

Paper \cite{breimanbag96} discusses, that this procedure is especially useful for learners
$bl$ which are unstable---small perturbations in the data have
big impact on the resulting model. Aggregating the bootstrapped models essentially
introduces robustness to such models. Examples of learners where the bagging is beneficial
are neural networks (where overfitting is often a serious problem) and 
regression trees (see Random Forests above).

\subsubsection{Stacking}
\label{mach:stack}
\begin{figure}[h]
\centering
\psfrag{L1}{1st Level}
\psfrag{L2}{2nd Level}
\psfrag{Input}{Input}
\psfrag{A}{\textbf{A}}
\psfrag{B}{\textbf{B}}
\psfrag{C}{\textbf{C}}
\psfrag{D}{\textbf{D}}
\psfrag{E}{\textbf{E}}

\includegraphics[trim={0 1.5cm 0 0},width=0.4\textwidth]{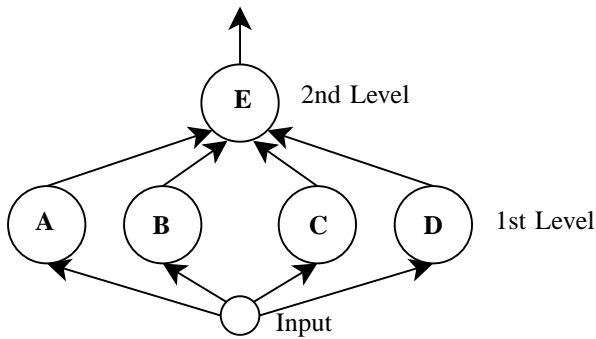}
\caption[Topology of the stacked ensemble]{The topology of the
    stacking ensemble method.
    \textbf{A}, \textbf{B}, \textbf{C} and \textbf{D} are the level 1 learners, \textbf{E} is
    the level 2 learner.
}
\label{fig:stacking}
\end{figure}

is a more sophisticated approach. The original idea was pioneered
by \cite{wolpert92} and extended by \cite{chan1996extensible,breiman96}.
The method is basically a two level hierarchical model of learners with
a clever scheme for training.
The first level is composed by an ensemble of learners.
The second level is a single learner which aggregates guesses from the 1st level models
and outputs the final prediction. Figure~\ref{fig:stacking} shows the topology.

The training dataset is divided into smaller parts (by cross-validation, see
Section~\ref{mach:crossval}). The 1st level learners are trained on some of them and
their \emph{generalization} biases are measured by testing their performance on 
the rest. The 2nd level learner learns to correct these---it learns what the correct output
is, given what the 1st level predictors output.
Algorithm~\ref{alg:stacking} describes the procedure in more detail.
Usually, the 2nd level learner is a simple linear regression, in this work we use
non-linear models as well.

Having diverse base learners (various models with different biases) often proves
to increase the prediction performance. The performance of stacking
is usually better than the best of the base learners on its own. It is not the case,
however, that having badly performing learners in the ensemble does not worsen the 
performance. Choosing the right set of 1st level learners is very important if we are to attain
the best performance, as is the choice of the 2nd level aggregating learner and the number
of folds for the cross-validation step. Usually, all these parameters are hand-tuned;
in this work, we use a genetic algorithm (Secion~\ref{mach:ga}).

\begin{algorithm}[]
\caption[Stacking]{Stacking for Regression}
\label{alg:stacking}

\SetKwData{AllPat}{AllPat}
\SetKwData{TopPat}{TopPat}
\SetKwData{Aa}{A}
\SetKwData{Na}{N}
\SetKwData{Ga}{G}
\SetKwFunction{CrossValidation}{CrossValidation}
\SetKwFunction{Compose}{Compose}

 \SetKwInOut{Input}{input}
 \SetKwInOut{Output}{output}

 \Input{an ordered set of 1st level learners $ensemble$,
        a level 2 learner $l2$,
        training data $Tr$,
        number of folds $Folds$}

 \Output{regression function $f$}
 \BlankLine

 \tcc{Training set for the level 2 learner.}
 $L2Tr \leftarrow \{\}$\;

 \ForEach{$(Tr', Ts')$ in \CrossValidation{$Tr$, $Folds$}}{
     \tcc{The level 1 learners trained on split $Tr'$.}
     $L1 \leftarrow ( ensemble_0 (Tr'), \ldots, ensemble_n (Tr') )$\;
     \ForEach{$(x',y')$ in $Ts'$}{

         \tcc{Responses of level 1 predictors to unseen $x'$ and the real reply $y'$.}
         $L2Tr \leftarrow L2Tr \cup \{(( L1_0(x'), \ldots, L1_n(x')), y')\}$\;
     }
     
 }

 \tcc{Train the level 1 learners on the real data.}
 $L1 \leftarrow ( ensemble_0 (Tr), \ldots, ensemble_n (Tr) )$\;
 \tcc{Train the level 2 learner on the prepared data.}
 $L2 \leftarrow l2(L2Tr)$\;

 \KwRet{\Compose{$L1$, $L2$}}\;
\end{algorithm}


\subsection{Evolving Stacking Ensembles}
\label{mach:ga}

We have discussed that ensemble learning might be beneficial in terms of performance.
For stacking, it is desirable to form the ensemble out of diverse base learners.
The problem however is how to choose the learners into the ensemble. This becomes
apparent once one tries to hand-tune the parameters of different base-learners,
find the best combination of them and find the best aggregating 2nd level learner.

We have used a simple genetic algorithm (\emph{GA}) to search the space of possible
ensembles for the stacking. Genetic algorithms are
an universal optimization tool, see~\cite{whitley94} for a
good tutorial. The general procedure is iterative. In each iteration,
individuals (candidate solutions) are evaluated using a \emph{fitness
function}, and an intermediate population is formed by randomly choosing
individuals, with probability proportional to the fitness (roulette selection).
From this
intermediate population, the population for the next step is taken by
making pairwise \emph{crossover} operation and mutation on the newly
formed individuals.

In the text below, we operate with a set of base learners $BL$, from which we
choose the learners into the ensemble. We should note that the set of base learners
$BL$ is not strictly limited to learners we have listed as
base in Section~\ref{mach:base_learners}---we use both differently
parameterized base learners and bagged neural networks.

We have used a very simple \textbf{encoding for an individual}.
An individual is a triple of $(I, Folds, \vec v)$.
The first two values $I$ and $Folds$ define the 2nd level learner. $I$ is 
the index of the 2nd level aggregating learner in $BL$ and
$Folds$ is the number of folds for the stacking procedure.
The vector $\vec v$ of size $|BL|$ marks a subset of $BL$ that forms the ensemble:
$\vec v_i = 1$ if the base learner $BL_i$ belongs to the ensemble; $\vec v_i = 0$
when it does not.

\SetKwFunction{MutateM}{MutateM}
\SetKwFunction{MutateV}{MutateV}

We have used two independent \textbf{mutations} to modify the individuals.
Firstly, with probability $Pm_M$,
we either change $I$ to any of $1 \ldots |BL|$, or we change
the number of $Folds$ to $2 \ldots 6$, (\MutateM in the pseudocode).
Whether we change $I$, or $Folds$ is decided using a further random coin toss.
Secondly, with probability $Pm_v$ a random position $i$ in $v$ is selected and
the bit $v_i$ is swapped, (\MutateV in the pseudocode).

The \textbf{crossover operation} of parents $P = (I, Folds, \vec v)$ and
$P' = (I', Folds', \vec v')$ selects a~random position $i \in \{1 \ldots |BL|\}$
and outputs the following tuple
$$(I, Folds, (v_1, \ldots, v_i, v'_{i+1}, \ldots, v'_{|BL|}))$$ as the new individual. Please
note that the index $I$ (and number of $Folds$)
of the 2nd level learner is taken from the first parent $P$.
This is compensated for by the fact that crossover is always performed in pairs
(lines~\ref{cross} -- \ref{cross2} in Algorithm~\ref{alg:ga}).

The \textbf{fitness} function we have used is inversely proportional to
$RMSE$ error of the resulting stacked ensemble.

Also, to make sure we do not lose the best solution, we have used elitism,
which brings the top $E$ individuals unchanged into the next generation.

\begin{algorithm}[]
\caption{Genetic Algorithm for finding optimal stacking ensemble}
\label{alg:ga}

\SetKwData{AllPat}{AllPat}
\SetKwData{TopPat}{TopPat}
\SetKwData{Aa}{A}
\SetKwData{Na}{N}
\SetKwData{Ga}{G}
\SetKwFunction{Xover}{Crossover}
\SetKwFunction{Best}{Best}
\SetKwFunction{Fitness}{Fitness}
\SetKwFunction{RouletteSelection}{RouletteSelection}
\SetKwFunction{RandomPopulation}{RandomPopulation}
\SetKwFunction{TakeTop}{TakeTop}
\SetKwFunction{Rnd}{Rnd}

 \SetKwInOut{Input}{input}
 \SetKwInOut{Output}{output}

 \Input{size of the population $S$,
        size of the elite $E$,
        probabilities of mutation $Pm_M$ and $Pm_v$,
        maximal number of steps $Max$}

 \Output{The best individual.}
 \BlankLine
 $Pop \leftarrow$ \RandomPopulation{$S$}\;
 \tcc{The best individual so far.}
 $Best \leftarrow \{\}$\;
 \ForEach{$iteration$ in $1 \ldots Max$}{
     $evaluation \leftarrow$ \Fitness{$Pop$}\;
     \tcc{$PI$ is the intermediate population.}
     $PI \leftarrow$ \RouletteSelection{$Pop$, $evaluation$}\;
     \tcc{$PN$ is the intermediate population after \Xover.}
     $PN \leftarrow \{\}$\;
     \ForEach{$i$ in $1 \ldots (S - E)/2$}{
         $PN \leftarrow PN$ $\cup$ \Xover{$PI[2*i]$, $PI[2*i+1]$} \;
        \label{cross}
         $PN \leftarrow PN$ $\cup$ \Xover{$PI[2*i+1]$, $PI[2*i]$} \;
        \label{cross2}
     }
     \tcc{Save the best individual.}
     $Best \leftarrow$ \TakeTop{$Pop$, $evaluation$, 1}\;
     \tcc{Top $E$ best continue unchanged.}
     $Pop \leftarrow$ \TakeTop{$Pop$, $evaluation$, $E$}\;

     \ForEach{$individual$ in $PN$}{
         \If{$\Rnd{0,1} < Pm_M$}{
             $individual \leftarrow $\MutateM{$individual$}\;
         }
         \If{$\Rnd{0,1} < Pm_v$}{
             $individual \leftarrow $\MutateV{$individual$}\;
         }
         
         $Pop \leftarrow Pop$ $\cup \{individual\}$\;
     }
     
 }

 \KwRet{$Best$}\;
\end{algorithm}

\section{Evaluating Learners}
\label{mach:cmp}
To compare performances of different regression functions (learners),
we need a~reliable metric.
The goal is to estimate the performance of a particular regression function
on real unseen data. We can estimate this performance by splitting the data
into parts that are only used for training ($Tr$) and testing ($Ts$).

\subsection{Cross-Validation}
\label{mach:crossval}
Cross-validation is a standard statistical technique for estimation of parameters.
The idea is to split the data into $k$ disjunct subsets (called \emph{folds}), and then
iteratively compose the training and testing sets and measure errors.
In each of the $k$ iterations, $k$-th fold is chosen as the testing data, and
all the remaining $k-1$ folds form the training data. The division into the folds is
done randomly, and so that the folds have approximately the
same size (in cases where the number of
samples $|D|$ is not divisible by $k$, some folds are slightly smaller than others).
Please note that each sample from the data is a part of the testing fold exactly once (it is
part of a training set $k-1$ times).
Refer to~\cite{crossval} for details.


\subsection{Error Analysis}
\label{mach:residuals}
A commonly used performance measure is the mean square error ($MSE$) which estimates variance of
the error distribution. We use its square root ($RMSE$) which is an estimate of
standard deviation of the predictions,
$$ RMSE = \sqrt{\frac{1}{|Ts|} \sum_{(ev, y) \in Ts}{ (predict(ev) - y)^2}},$$
where the machine learning model $predict$ is trained on the
training data $Tr$ and $Ts$ denotes the testing data.

\section{Experiments and Results}
\label{sec:expe}

\subsection{Strength}
One of the two major domains we have tested our framework on is the prediction of player
strength. In the game of Go, strength (amateur) is measured by kyu (student) and dan
(master) ranks.  The kyu ranks decrease from about 20th kyu (absolute beginner) to
1st kyu (fairly strong player), the scale continues by dan ranks, 1 dan
(somewhat stronger than 1 kyu), to 6 dan (a very strong player).

\subsubsection*{Dataset}

We have collected a large sample of games from the public archives of
the Kiseido Go server~\cite{KGSArchives}, the sample consists of over
100 000 records of games. The records were divided by player's strength
and preprocessed as is detailed in~\cite{Moudrik15}. Here, it is enough to
note that the dataset consists of 120 independent pairs $(x, y)$ for each
of the 26 ranks, where $x$ is the feature vector
described in~\cite{Moudrik15} (dimension of $x$ was 1040), and $y$ is
the target variable, which is one number describing the rank (1-26).

\subsubsection*{Results}

\label{exp:str_regr}

In the process of finding the best learner, we started with a hand-tuned learner
shown in Table~\ref{tab:hand_tuned}. Using this learner (which we found to perform
reasonably well, as shown in Table~\ref{tab:str_reg_res})
we evaluated different feature extractors (previous section).
At first, the dataset was processed using the best feature
extractors, which were concatenated
to form the data $T$ for regression.

We then used the genetic algorithm (Section~\ref{mach:ga}; abbreviated to \emph{GA})
to find the best performing
stacked ensemble. The initial population was seeded by the hand tuned learner.
The parameters of the genetic algorithm are given in Table~\ref{tab:ga}.

\begin{table}
\begin{center}
\caption{The parameters of the genetic algorithm
for the strength dataset.}
\label{tab:ga}
\begin{tabular}{|c|c|}
\hline
\textbf{Parameter} & \textbf{Value} \\
\hline
Set of base learners $BL$ & Is given in Table~\ref{tab:base_learners}. \\
Population size $S$ & 16 \\
Elite size $E$ & 1 \\
Number of iterations $Max$ & 100 \\
Mutation probability $Pm_I$ & 0.2 \\
Mutation probability $Pm_v$ & 0.5 \\
Fitness function & $1/RMSE$ of the resulting\\
				 & stacked learner The $RMSE$\\
				&is computed using 5-fold cross-validation. \\
\hline
\end{tabular}
\end{center}
\end{table}

\begin{table}
\begin{center}
\caption{Strength: Best GA Stacking Ensemble.}
\label{tab:str_best}
\begin{tabular}{|c|c|}
\hline
\textbf{Ensemble l.} & \textbf{Settings} \\
\hline

Stacking & 6 folds, level 2 learner: Bagged ($20\times$) NN:\\
& $\epsilon = 0.005$, $max = 500$ iter., 1 layer , 10 neurons. \\ \hline

\textbf{Base l.} & \textbf{Settings} \\
\hline
Mean regression & --- \\
PLS regression & $l = 3$ \\
Random Forests & $N = 50$\\
Neural network & $\epsilon = 0.001$, $max = 200$ iter., 1 layer, 20 neurons. \\

$k$-nn & $k = 20$, $\alpha = 20$, $\delta = \text{Euclidean}$. \\
$k$-nn & $k = 40$, $\alpha = 10$, $\delta = \text{Manhattan, Euclidean}$. \\
$k$-nn & $k = 40$, $\alpha = 20$, $\delta = \text{Euclidean}$. \\
$k$-nn & $k = 50$, $\alpha = 10$, $\delta = \text{Manhattan}$. \\
$k$-nn & $k = 50$, $\alpha = 20$, $\delta = \text{Manhattan, Euclidean}$. \\
$k$-nn & $k = 60$, $\alpha = 10$, $\delta = \text{Euclidean}$. \\
$k$-nn & $k = 60$, $\alpha = 20$, $\delta = \text{Euclidean}$. \\

Bagged NN & 20 $\times$ NN:
$\epsilon = 0.001$, $max = 100$, 1 layer, 10 neur. \\
Bagged NN & 40 $\times$ NN:
$\epsilon = 0.005$, $max = 100$, 1 layer, 10 neur. \\
Bagged NN & 40 $\times$ NN:
$\epsilon = 0.001$, $max = 500$, 1 layer, 20 neur. \\
Bagged NN & 20 $\times$ NN:
$\epsilon = 0.005$, $max = 200$, 1 layer, 20 neur. \\
Bagged NN & 40 $\times$ NN:
$\epsilon = 0.005$, $max = 500$, 1 layer, 20 neur. \\

\hline
\end{tabular}
\end{center}
\end{table}

\begin{table}
\begin{center}
\caption{The hand-tuned learner.}
\label{tab:hand_tuned}
\begin{tabular}{|c|c|}
\hline
\textbf{Ensemble learner} & \textbf{Settings} \\
\hline

Stacking & 4 folds, level 2 learner:
NN, $\epsilon = 0.005$,\\
&  $max = 100$ iter., 1 layer, 10 neurons. \\ \hline

\textbf{Base learners} & \textbf{Settings} \\
\hline
Mean regression & --- \\
PLS regression & $l = 3$ \\

$k$-NN& $k = 50$, $\alpha = 20$, $\delta = \text{Manhattan}$. \\

Random Forests & $N = 50$\\

Bagged NN & 20 $\times$ NN: $\epsilon = 0.001$, $max = 100$ iter.,\\
& 1 layer, 10 neurons. \\

\hline
\end{tabular}
\end{center}
\end{table}

Unfortunately, the time needed for a~single
iteration was very large in this setting (see Discussion for more
detailed treatment of the time complexity).
To speed up the process, we used a sub-sampled dataset for computing
the fitness during the GA (by randomly taking $1/10$ prior
to the running of the GA). We assume, that this is not a principal
obstacle for finding the best learner, since the down-sampling
should degrade the performance of the learners
\emph{systematically}---the ordering of fitnesses is expected to be
more or less the same, though the fitness values surely differ.
The run of genetic algorithm took on average approximately 2.5 hours per
iteration on a 4-core commodity laptop in this setting.

\begin{figure}
\centering
\psfrag{Meta}{Level-2 Learner}
\psfrag{Base}{Base Learners}
\psfrag{M}{\textbf{Mean}}
\psfrag{P}{\textbf{PLS}}
\psfrag{RF}{\textbf{RF}}
\psfrag{rf}{50}
\psfrag{NN}{\textbf{NN}}
\psfrag{NN1}{\textbf{NN}}
\psfrag{NN2}{\textbf{NN}}
\psfrag{NNm}{\textbf{NN}}
\psfrag{mn}{} 
\psfrag{kn1}{$k$-nn}
\psfrag{kn2}{$k$-nn}
\psfrag{bag}{Bag}
\psfrag{t1}{$9\times$}
\psfrag{t2}{$5\times$}
\includegraphics[width=0.5\textwidth]{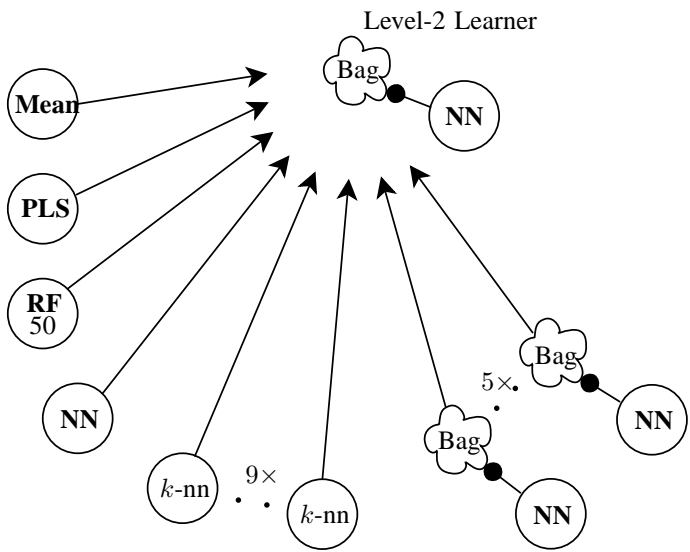}
\caption{
    Structure of the best stacking ensemble found by the
    Genetic Algorithm. The circle marks a normal learner,
    with description within, the ``cloud'' denotes a bagging learner.
    The corresponding bagged learner is
    connected using the circle-ended arrow. Precise descriptions of the learners are given
    in Table~\ref{tab:str_best}. \textbf{Mean} is the Mean regression, \textbf{PLS} Partial least squares
    regression, \textbf{RF} Random Forests, \textbf{NN} various neural networks and $k$-nn
    is obviously the $k$-nearest neighbor learner.
  }
\label{fig:ga_str}
\end{figure}

\begin{figure*}
\vspace{0cm}
\centering
\psfrag{Iteration}{Iteration}
\psfrag{Fitness}{$\mathbf{RMSE}$}
\includegraphics[width=\textwidth,height=7.8cm]{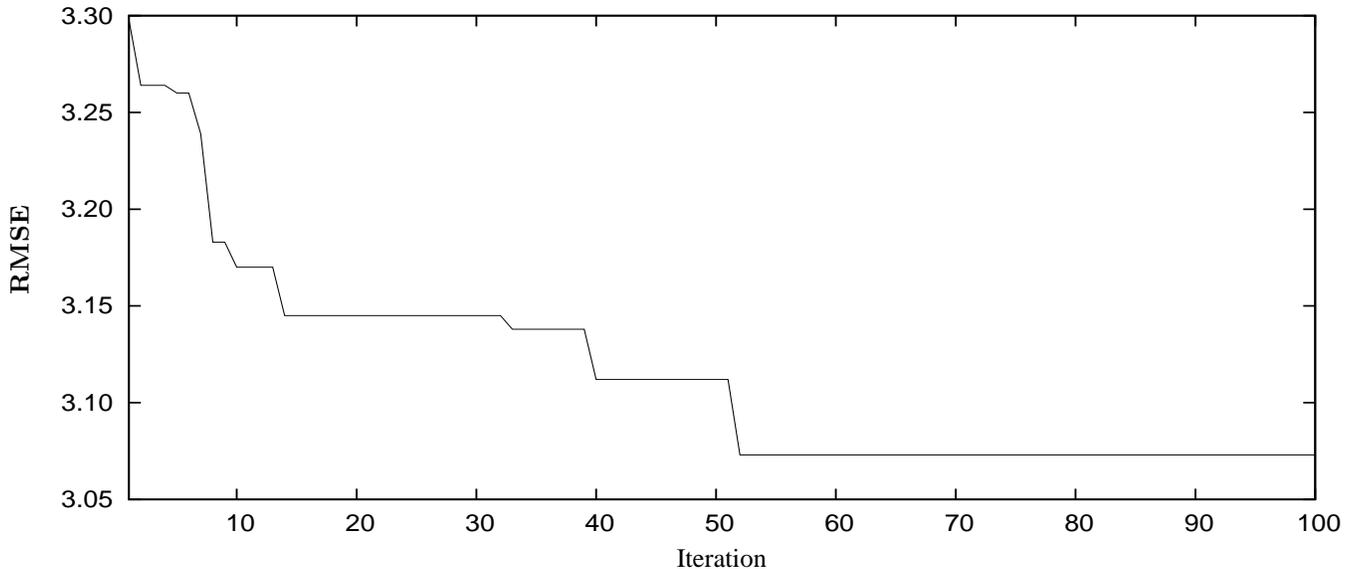}
\caption[Evolution of $RMSE$ error in time (strength)]{
    Evolution of $\mathbf{RMSE}$ error during the run of the genetic algorithm for 
	finding an optimal stacking ensemble for the (sub-sampled) strength data.
  }
\label{fig:str_ga}
\end{figure*}

The performance of the best ensemble found by the GA (on the full dataset)
are given in Table~\ref{tab:str_reg_res} along with other learners to compare performances.
The resulting learner (Table~\ref{tab:str_best}) is fairly complex
as Figure~\ref{fig:ga_str} shows. Evolution of the $\mathbf{RMSE}$
error in time is given in Figure~\ref{fig:str_ga}.

\begin{table}
\begin{center}
\caption{Regression performance of different learners on the full
    dataset. The results were computed using 5-fold cross-validation.
Parameters of the best GA stacking ensemble
    are given in Table~\ref{tab:str_best}, the other learners are taken from
	the Initial hand-tuned learner from Table~\ref{tab:hand_tuned}.} 
\begin{tabular}{|c|c|c|}
\hline
\textbf{Learner} & $\mathbf{RMSE}$ & \textbf{Mean cmp}\\

\hline
Mean regression & 7.507 & 1.00 \\ \hline
Random Forrest& 3.869 & $1.94$\\
PLS & 3.176 & $2.36$\\
Bagged NN & 2.66 & $2.82$\\
Hand-tuned learner & 2.635 & $2.85$\\\hline
Best GA stacking ensemble & 2.607 & $2.88$\\
\hline
\end{tabular}
\end{center}
\label{tab:str_reg_res}
\end{table}

\subsection{Style}
\label{exp:style}
Apart from the strength estimation, we also tested the framework presented in this work
to test prediction of playing styles of professional players. Playing style has different
aspects, some of them are vaguely defined (for details, see~\cite{GoGoD:styles}).
Moreover, none of them has
clear definition in a mathematical sense. To capture these notions at least approximately, we
came up with four axes, whose ends correspond to opposing principles in traditional 
Go knowledge. Next, we used a questionnaire (submitted to domain experts) to find
the style values for a set of well known professional players from the 20th century.

\begin{center}
\begin{tabular}{|c|c|c|}
\hline
\textbf{Style} & \textbf{1} & \textbf{10}\\ \hline
Territoriality & Moyo & Territory \\
Orthodoxity & Classic & Novel \\
Aggressivity& Calm & Fighting \\
Thickness & Safe & Shinogi \\ \hline
\end{tabular}
\end{center}

\subsubsection*{Dataset}
\label{exp:style_dataset}

The collection of games
in this dataset comes from the Games of Go on Disk (\emph{GoGoD}) database by \cite{GoGoD}.
This database contains more than 70 000 games, spanning from the ancient times
to the present.  We chose a small subset of well known players (mainly from the 20th century)
and asked some experts (professional and strong amateur players) to evaluate
these players using a questionnaire. \cite{GoStyleArxiv}
The experts (Alexander Dinerchtein 3-pro, Motoki Noguchi 7-dan,
Vladim\'{i}r Dan\v{e}k 5-dan and V\'{i}t Brunner 4-dan) were asked to value 
the players on four scales, each ranging from 1 to 10.

For each of 24 professional players, we obtained 12 pairs $(x, y)$, where 
$x$ is the feature vector obtained from the games as described in~\cite{Moudrik15}
(dimension of $x$ was 640),
and $y$ is one of the 4 styles---basically, we view the problem as 4 different
regression problems which share the same feature vectors.

\subsubsection*{Results}
\label{exp:style_regr}

We used the genetic algorithm to determine the best ensemble learner. 
During the process, we have encountered over-fitting problems concerning the
very small size of the dataset.

At first, we chose the parameters of the GA to be the same as in the problem of
strength (Table~\ref{tab:ga}), with the exception of the fitness
function.  The $RMSE$ error was computed in the same manner as in the style feature
extraction (one learner for all the styles, the fitness of a learner is average $RMSE$
on the different styles).
Similarly to the case of strength, it turned out that it was not possible to use cross-validation
on the full dataset because of time constraints. We tried to workaround this by
subsampling the data prior to the experiment, but due to very small size of
the dataset, this resulted in over-fitting of the resulting ensemble model.

Secondly, we tried not to use the cross-validation, but to use proportional division
scheme instead---the fitness is evaluated by randomly taking 70\% of the dataset
for training and the rest for testing; in each of the iterations, this is done anew
to mitigate any effects caused by biased random split (dividing the dataset once prior
to the run would cause over-fitting). Unfortunately, this too did not yield satisfactory
results. Even though over-fitting was not the case, the genetic algorithm was not able
to consistently improve the ensembles---the sub-sampled datasets in each of the iterations
were too different to ensure that the best individuals from one iteration would have
good chances in the next one. This rendered the genetic algorithm unsuccessful.

Consequently we concluded, that the robust cross-validation with the full dataset is
necessary and that we thus need to compensate for the increased resource consumption
differently. We did this by limiting the population size to 10 individuals and 
most importantly, by limiting the ensemble to contain at most 5 base learners. Technically,
this is done by randomly removing excess number of base learners from each individual
at the end of each iteration. The parameters of the final genetic algorithm are listed
in Table~\ref{tab:sty_ga_param}.

\begin{figure*}
\vspace{0cm}
\centering
\psfrag{Iteration}{Iteration}
\psfrag{Fitness}{$\mathbf{RMSE}$}
\psfrag{territoriality}{\hspace{6ex}\small Territoriality}
\psfrag{orthodoxity}{\hspace{5.5ex}\small Orthodoxity}
\psfrag{aggressivity}{\hspace{5ex}\small Aggressivity}
\psfrag{thickness}{\hspace{4ex}\small Thickness}
\includegraphics[width=\textwidth,height=7cm]{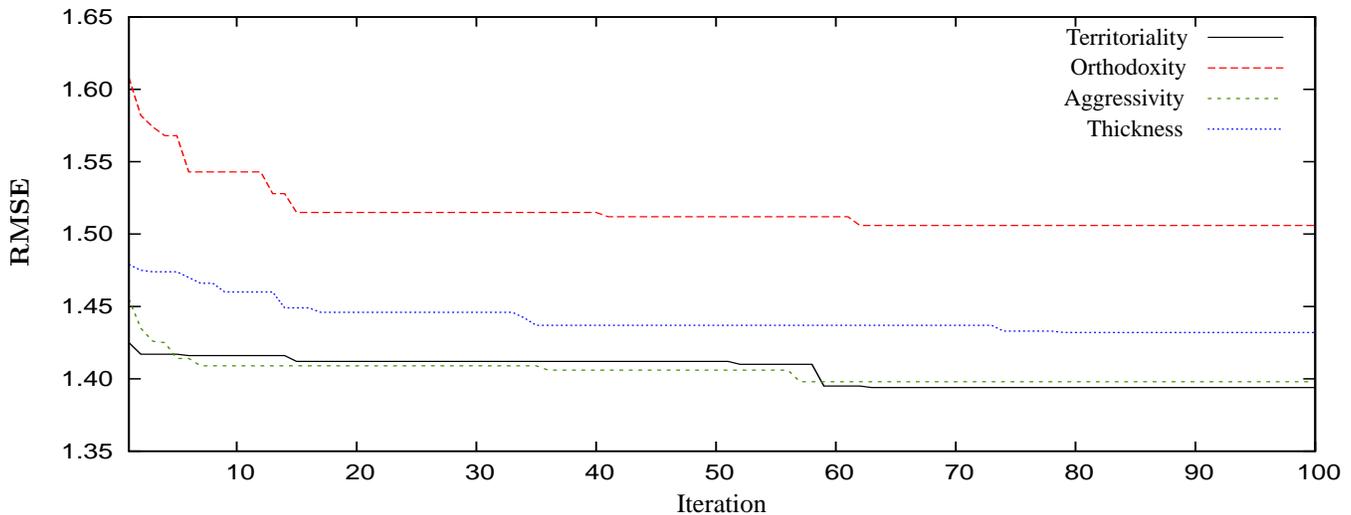}
\caption[Evolution of $RMSE$ error in time (styles)]{
    Evolution of $\mathbf{RMSE}$ error during the run of the genetic algorithm for 
    finding an optimal stacking ensemble for the style data.
  }
\label{fig:sty_ga}
\end{figure*}

The performances of the best learners found are given in Table~\ref{tab:sty_reg_res}.
The parameters of the models are omitted for brevity, see \cite{Moudrik13} for more.
Development of the $\mathbf{RMSE}$ error in time is given in Figure~\ref{fig:sty_ga}.
Each run of the genetic algorithm (for different styles) took approximately one
hour of CPU time per iteration.

\begin{table}[h]
\begin{center}
\caption[Parameters for the genetic algorithm (styles)]{The parameters of the genetic algorithm
for the style dataset.}
\label{tab:sty_ga_param}
\begin{tabular}{|c|c|}
\hline
\textbf{Parameter} & \textbf{Value} \\
\hline
Set of base learners $BL$ & Is given in Table~\ref{tab:base_learners}. \\
Population size $S$ & 10 \\
Elite size $E$ & 1 \\
Number of iterations $Max$ & 100 \\
Mutation probability $Pm_I$ & 0.2 \\
Mutation probability $Pm_v$ & 0.5 \\
Ensemble size limit & 5 \\
Fitness function & $1/RMSE$ of the resulting stacked\\
				 & learner. The $RMSE$ is computed\\
				 & using 5-fold cross-validation. \\
\hline
\end{tabular}
\end{center}
\end{table}

\begin{table}[h]
\begin{center}
\caption[Regression performance of different learners (styles)]{Regression performance of different learners on the full dataset. The results were computed using 5-fold cross-validation.}
\label{tab:sty_reg_res}
\begin{tabular}{|c|c|c|}
\hline
&
\multicolumn{2}{|c|}{ $\mathbf{RMSE}$ } \\
\hline
\textbf{Learner} & Territoriality & Orthodoxity\\

\hline
Mean regression & 2.403 & 2.421\\
Hand tuned learner & 1.434 & 1.636\\
The best GA learner & 1.394 & 1.506\\

\hline
\textbf{Learner}  & Aggressivity & Thickness \\
\hline
Mean regression  & 2.179 & 1.682 \\
Hand tuned learner & 1.423 & 1.484 \\ 
The best GA learner & 1.398 & 1.432 \\
\hline
\end{tabular}
\end{center}
\end{table}

\section{Discussion}
\label{sec:disc}

The results in both domains show that evolving stacking ensembles non-trivially
improves upon the performance of the best hand tunes ensemble, circa by 1.5\% for
the case of strength and by 4\% for the style domains (averaged over different styles).
The resulting ensembles also show a lot of diversity.

The main drawback of the methods described
is clearly the time consumed, which---even for such relatively small datasets---is
in orders of hours per iteration. The main cause of this is the cross-validation
performed at various levels, multiplying time complexity. At the outer level, it is
run to obtain better error estimates. The cross-validation is also used in the inner
loop during training Stacking ensembles, and moreover, in some of the base learners.
In this extreme case, the complexity of training the base model is multiplied by
number of $Folds^3$. With some effort, the large sequential time could also be exchanged
for large number of machines, since it is easy to train different folds in parallel.
Furthermore, the time-complexity can be lowered by limiting the number of learners in
the ensemble, or by restricting the base learners used by the stacking ensemble to
be simple and fast models (with possible loss of precision).

The prediction of player attributes as demonstrated in this work has been
(together with the feature extraction presented in \cite{Moudrik15})
combined in an online web application\footnote{\url{http://gostyle.j2m.cz}},
which evaluates games submitted by players and predicts their playing strength
and style. The predicted strength
is then used to recommend relevant literature and the playing style is
utilized by recommending relevant professional players to review.
So far, the web application has served thousands of players and it was generally
very well received.
We are aware of only two tools, that do something
alike, both of them are however based on a predefined questionnaire. The first one
is the tool of~\cite{style:baduk}---the user answers 15 questions and based 
on the answers he gets one of predefined recommendations.
The second tool is not available at
the time of writing, but the discussion at~\cite{senseis:which_pro}
suggests, that it computed distances to some professional players based on user's
answers to 20 questions regarding the style. We believe that our approach
is more precise, both because it takes into account many different
aspects of the games~\cite{Moudrik15}, and because the methods presented
in this paper are able to use the information well.

\section{Conclusion}
\label{sec:conc}

The paper presents machine-learning algorithm
which evolves non-linear stacking ensembles of learners.
The algorithm is applied on two computer Go domains,
prediction of player's strength and different aspects of his style.
In both these domains, the algorithm outperforms other methods,
with the disadvantage of taking relatively large amount of time;
solutions to this problem are proposed.

\section{Implementation}
\label{sec:impl}

The code used in this work is released online as a part of
GoStyle project~\cite{GoStyleWeb}.  The majority of the source code
is implemented in the Python programming language~\cite{Python27}.

The machine learning models were implemented and evaluated using the
Orange Datamining suite~\cite{demsar13a} and the Fast Artificial Neural Network
library FANN~\cite{Nissen2003}.  We used the Pachi Go engine~\cite{Pachi}
for the raw game processing.

\subsection*{Acknowledgment} 
This research has been partially supported by the Czech Science Foundation
project no.~15-18108S. J. Moud\v{r}\'\i k has been supported by the
Charles University Grant Agency project no.~364015 and by SVV project
no.~260 224.

\bibliographystyle{IEEEtran}
\bibliography{clanek}

\begin{thebibliography}{10}
\providecommand{\url}[1]{#1}
\csname url@samestyle\endcsname
\providecommand{\newblock}{\relax}
\providecommand{\bibinfo}[2]{#2}
\providecommand{\BIBentrySTDinterwordspacing}{\spaceskip=0pt\relax}
\providecommand{\BIBentryALTinterwordstretchfactor}{4}
\providecommand{\BIBentryALTinterwordspacing}{\spaceskip=\fontdimen2\font plus
\BIBentryALTinterwordstretchfactor\fontdimen3\font minus
  \fontdimen4\font\relax}
\providecommand{\BIBforeignlanguage}[2]{{%
\expandafter\ifx\csname l@#1\endcsname\relax
\typeout{** WARNING: IEEEtran.bst: No hyphenation pattern has been}%
\typeout{** loaded for the language `#1'. Using the pattern for}%
\typeout{** the default language instead.}%
\else
\language=\csname l@#1\endcsname
\fi
#2}}
\providecommand{\BIBdecl}{\relax}
\BIBdecl

\bibitem{GellySilver2008}
S.~Gelly and D.~Silver, ``Achieving master level play in 9x9 computer go,'' in
  \emph{AAAI'08: Proceedings of the 23rd national conference on Artificial
  intelligence}.\hskip 1em plus 0.5em minus 0.4em\relax AAAI Press, 2008, pp.
  1537--1540.

\bibitem{Moudrik15}
J.~Moud{\v{r}\'{i}}k, P.~Baudi{\v{s}}, and R.~Neruda, ``Evaluating go game
  records for prediction of player attributes,'' in \emph{IEEE Computational
  Intelligence in Games 2015}.\hskip 1em plus 0.5em minus 0.4em\relax IEEE,
  2015, pp. 162--168, in Print.

\bibitem{Moudrik13}
\BIBentryALTinterwordspacing
J.~Moud{\v{r}\'{i}}k, ``Meta-learning methods for analyzing go playing
  trends,'' Master's thesis, Charles University, Faculty of Mathematics and
  Physics, Prague, Czech Republic, 2013. [Online]. Available:
  \url{http://www.j2m.cz/~jm/master_thesis.pdf}
\BIBentrySTDinterwordspacing

\bibitem{hansen1990neural}
L.~K. Hansen and P.~Salamon, ``Neural network ensembles,'' \emph{IEEE
  Transactions on Pattern Analysis \& Machine Intelligence}, no.~10, pp.
  993--1001, 1990.

\bibitem{breimanbag96}
\BIBentryALTinterwordspacing
L.~Breiman, ``Bagging predictors,'' \emph{Mach. Learn.}, vol.~24, no.~2, pp.
  123--140, Aug. 1996. [Online]. Available:
  \url{http://dx.doi.org/10.1023/A:1018054314350}
\BIBentrySTDinterwordspacing

\bibitem{breiman01}
------, ``Random forests,'' \emph{Machine Learning}, vol.~45, no.~1, pp. 5--32,
  Oct. 2001.

\bibitem{kittler1998combining}
J.~Kittler, M.~Hatef, R.~P. Duin, and J.~Matas, ``On combining classifiers,''
  \emph{Pattern Analysis and Machine Intelligence, IEEE Transactions on},
  vol.~20, no.~3, pp. 226--239, 1998.

\bibitem{hinton2012improving}
G.~E. Hinton, N.~Srivastava, A.~Krizhevsky, I.~Sutskever, and R.~R.
  Salakhutdinov, ``Improving neural networks by preventing co-adaptation of
  feature detectors,'' \emph{arXiv preprint arXiv:1207.0580}, 2012.

\bibitem{boosting}
\BIBentryALTinterwordspacing
Y.~Freund and R.~Schapire, ``A desicion-theoretic generalization of on-line
  learning and an application to boosting,'' in \emph{Computational Learning
  Theory}, ser. Lecture Notes in Computer Science, vol. 904.\hskip 1em plus
  0.5em minus 0.4em\relax Springer Berlin Heidelberg, 1995, pp. 23--37.
  [Online]. Available: \url{http://dx.doi.org/10.1007/3-540-59119-2_166}
\BIBentrySTDinterwordspacing

\bibitem{wolpert92}
\BIBentryALTinterwordspacing
D.~H. Wolpert, ``Stacked generalization,'' \emph{Neural Networks}, vol.~5, pp.
  241--259, 1992. [Online]. Available:
  \url{http://www.machine-learning.martinsewell.com/ensembles/stacking/Wolpert1992.pdf}
\BIBentrySTDinterwordspacing

\bibitem{ting1999issues}
K.~M. Ting and I.~H. Witten, ``Issues in stacked generalization,'' 1999.

\bibitem{chan1996extensible}
P.~K.-W. Chan, ``An extensible meta-learning approach for scalable and accurate
  inductive learning,'' Ph.D. dissertation, Columbia University, 1996.

\bibitem{breiman96}
\BIBentryALTinterwordspacing
L.~Breiman, ``\BIBforeignlanguage{English}{Stacked regressions},''
  \emph{\BIBforeignlanguage{English}{Machine Learning}}, vol.~24, pp. 49--64,
  1996. [Online]. Available: \url{http://dx.doi.org/10.1007/BF00117832}
\BIBentrySTDinterwordspacing

\bibitem{style:baduk}
\BIBentryALTinterwordspacing
A.~Dinerchtein. (2012) What is your playing style? [Online]. Available:
  \url{http://style.baduk.com}
\BIBentrySTDinterwordspacing

\bibitem{senseis:which_pro}
\BIBentryALTinterwordspacing
{Sensei's Library}. (2013) Which pro do you most play like. [Online].
  Available: \url{http://senseis.xmp.net/?WhichProDoYouMostPlayLike}
\BIBentrySTDinterwordspacing

\bibitem{haykin_nn}
\BIBentryALTinterwordspacing
S.~Haykin, \emph{Neural Networks: A Comprehensive Foundation (2nd Edition)},
  2nd~ed.\hskip 1em plus 0.5em minus 0.4em\relax Prentice Hall, jul 1998.
  [Online]. Available: \url{http://www.worldcat.org/isbn/0132733501}
\BIBentrySTDinterwordspacing

\bibitem{Riedmiller1993}
M.~Riedmiller and H.~Braun, ``{A Direct Adaptive Method for Faster
  Backpropagation Learning: The RPROP Algorithm},'' in \emph{IEEE International
  Conference on Neural Networks}, 1993, pp. 586--591.

\bibitem{CoverHart1967}
T.~M. Cover and P.~E. Hart, ``Nearest neighbor pattern classification,''
  \emph{IEEE Transactions on Information Theory}, vol.~13, no.~1, pp. 21--27,
  1967.

\bibitem{pls}
\BIBentryALTinterwordspacing
R.~Rosipal and N.~Krämer, ``Overview and recent advances in partial least
  squares,'' in \emph{in ‘Subspace, Latent Structure and Feature Selection
  Techniques’, Lecture Notes in Computer Science}.\hskip 1em plus 0.5em minus
  0.4em\relax Springer, 2006, pp. 34--51. [Online]. Available:
  \url{http://citeseerx.ist.psu.edu/viewdoc/summary?doi=10.1.1.85.7735}
\BIBentrySTDinterwordspacing

\bibitem{breiman84}
L.~Breiman, J.~H. Friedman, O.~R. A., and C.~J. Stone, \emph{Classification and
  regression trees}.\hskip 1em plus 0.5em minus 0.4em\relax Monterey, CA:
  Wadsworth \& Brooks/Cole Advanced Books \& Software, 1984.

\bibitem{whitley94}
D.~Whitley, ``A genetic algorithm tutorial,'' \emph{Statistics and computing},
  vol.~4, no.~2, pp. 65--85, 1994.

\bibitem{crossval}
R.~Kohavi, ``A study of cross-validation and bootstrap for accuracy estimation
  and model selection.''\hskip 1em plus 0.5em minus 0.4em\relax Morgan
  Kaufmann, 1995, pp. 1137--1143.

\bibitem{KGSArchives}
\BIBentryALTinterwordspacing
W.~Shubert. (2013) {KGS} archives --- kiseido go server. [Online]. Available:
  \url{http://www.gokgs.com/archives.jsp}
\BIBentrySTDinterwordspacing

\bibitem{GoGoD:styles}
\BIBentryALTinterwordspacing
J.~Fairbairn. (winter 2011) {G}ames of {G}o on {D}isk --- {GoGoD}
  {E}ncyclopaedia and {D}atabase, {G}o players' styles. [Online]. Available:
  \url{http://www.gogod.co.uk/}
\BIBentrySTDinterwordspacing

\bibitem{GoGoD}
\BIBentryALTinterwordspacing
T.~M. Hall and J.~Fairbairn. (winter 2011) {G}ames of {G}o on {D}isk ---
  {GoGoD} {E}ncyclopaedia and {D}atabase. [Online]. Available:
  \url{http://www.gogod.co.uk/}
\BIBentrySTDinterwordspacing

\bibitem{GoStyleArxiv}
\BIBentryALTinterwordspacing
P.~Baudi{\v{s}} and J.~Moud{\v{r}\'{i}}k, ``On move pattern trends in a large
  go games corpus,'' \emph{Arxiv, CoRR}, October 2012. [Online]. Available:
  \url{http://arxiv.org/abs/1209.5251}
\BIBentrySTDinterwordspacing

\bibitem{GoStyleWeb}
\BIBentryALTinterwordspacing
J.~Moud{\v{r}\'{i}}k and P.~{B}audi{\v{s}}. (2013) {GoStyle} --- {D}etermine
  playing style in the game of {G}o. [Online]. Available:
  \url{http://gostyle.j2m.cz/}
\BIBentrySTDinterwordspacing

\bibitem{Python27}
\BIBentryALTinterwordspacing
{Python Software Foundation}. (2008, November) {P}ython 2.7. [Online].
  Available: \url{http://www.python.org/dev/peps/pep-0373/}
\BIBentrySTDinterwordspacing

\bibitem{demsar13a}
\BIBentryALTinterwordspacing
J.~Dem\v{s}ar \emph{et~al.}, ``Orange: Data mining toolbox in python,''
  \emph{Journal of Machine Learning Research}, vol.~14, pp. 2349--2353, 2013.
  [Online]. Available: \url{http://jmlr.org/papers/v14/demsar13a.html}
\BIBentrySTDinterwordspacing

\bibitem{Nissen2003}
S.~Nissen, ``Implementation of a fast artificial neural network library
  (fann),'' Department of Computer Science University of Copenhagen (DIKU),
  Tech. Rep., 2003, http://fann.sf.net.

\bibitem{Pachi}
\BIBentryALTinterwordspacing
P.~{B}audi{\v{s}} \emph{et~al.} (2012) Pachi --- {S}imple {G}o/{B}aduk/{W}eiqi
  {B}ot. [Online]. Available: \url{http://repo.or.cz/w/pachi.git}
\BIBentrySTDinterwordspacing

\end{thebibliography}

\end{document}